\def\year{2022}\relax
\documentclass[letterpaper]{article} % DO NOT CHANGE THIS
\usepackage{aaai22}  % DO NOT CHANGE THIS
\usepackage{times}  % DO NOT CHANGE THIS
\usepackage{helvet}  % DO NOT CHANGE THIS
\usepackage{courier}  % DO NOT CHANGE THIS
\usepackage[hyphens]{url}  % DO NOT CHANGE THIS
\usepackage{graphicx} % DO NOT CHANGE THIS
\usepackage{natbib}  % DO NOT CHANGE THIS AND DO NOT ADD ANY OPTIONS TO IT
\usepackage{caption} % DO NOT CHANGE THIS AND DO NOT ADD ANY OPTIONS TO IT
\DeclareCaptionStyle{ruled}{labelfont=normalfont,labelsep=colon,strut=off} % DO NOT CHANGE THIS
\usepackage{algorithm}
\usepackage{algorithmic}
\usepackage{booktabs}
\usepackage{newfloat}
\usepackage{listings}
\floatstyle{ruled}
\newfloat{listing}{tb}{lst}{}
\floatname{listing}{Listing}
\newcommand{\eg}{\textit{e}.\textit{g}.}
\newcommand{\etal}{\textit{et al}.}
\usepackage{subfigure}
\usepackage{multirow}

\usepackage{xcolor}
\title{Boosting the Transferability of Video Adversarial Examples \\
via Temporal Translation}
\author{
    Zhipeng Wei\textsuperscript{\rm 1,2}, 
    Jingjing Chen\textsuperscript{\rm 1,2}\equalcontrib,
    Zuxuan Wu\textsuperscript{\rm 1,2},
    Yu-Gang Jiang\textsuperscript{\rm 1,2}\equalcontrib
}
\affiliations{
    \textsuperscript{\rm 1}Shanghai Key Lab of Intelligent Information Processing, School of Computer Science, Fudan University \\
    \textsuperscript{\rm 2}Shanghai Collaborative Innovation Center on Intelligent Visual Computing
    
    zpwei21@m.fudan.edu.cn, \{chenjingjing, zxwu, ygj\}@fudan.edu.cn\\
}

\usepackage{bibentry}
\begin{document}

\maketitle

\begin{abstract}
Although deep-learning based video recognition models have achieved remarkable success, they are vulnerable to adversarial examples that are generated by adding human-imperceptible perturbations on clean video samples. As indicated in recent studies, adversarial examples are transferable, which makes it feasible for black-box attacks in real-world applications. Nevertheless, most existing adversarial attack methods have poor transferability when attacking other video models and transfer-based attacks on video models are still unexplored. To this end, we propose to boost the transferability of video adversarial examples for black-box attacks on video recognition models.
Through extensive analysis, we discover that different video recognition models rely on different discriminative temporal patterns, leading to the poor transferability of video adversarial examples. This motivates us to introduce a temporal translation attack method, which optimizes the adversarial perturbations over a set of temporal translated video clips.
By generating adversarial examples over translated videos, the resulting adversarial examples are less sensitive to temporal patterns existed in the white-box model being attacked and thus can be better transferred.
Extensive experiments on the Kinetics-400 dataset and the UCF-101 dataset demonstrate that our method can significantly boost the transferability of video adversarial examples. For transfer-based attack against video recognition models, it achieves a 61.56\% average attack success rate on the Kinetics-400 and 48.60\% on the UCF-101. Code is available at \url{https://github.com/zhipeng-wei/TT}.
\end{abstract}

%%%%%%%%% BODY TEXT
\section{Introduction}
Recent years have witnessed the great success of deep learning techniques on a series of tasks \cite{he2016deep, liu2018toward, feng2021empowering}, such as image recognition \cite{he2016deep, liu2020hyperbolic, chen2020study, chen2020zero}, Image segmentation \cite{jiao2021two}, object detection \cite{ren2016faster}, video recognition and retrieval \cite{wu2020dynamic, song2021spatial}. Therefore, DNNs have been widely applied in real-world applications, \eg, online recognition services, navigation robots, autonomous driving \cite{tian2018deeptest}, \textit{etc}. 
However, recent studies have identified that DNNs are vulnerable to adversarial examples \cite{goodfellow2014explaining, szegedy2013intriguing}, which are carefully crafted to fool DNNs by adding small human-imperceptible perturbations on clean samples. The existence of adversarial examples incurs security concerns in real-world applications. Thus, it has raised increasing attention over recent years. 

According to threat models, adversarial attacks can be divided into two categories: white-box attacks \cite{wei2019sparse} and black-box attacks \cite{guo2021meaningful}. In the white-box setting, an adversary can fully control and access DNN models, including model architectures as well as parameters. While in the black-box setting, the adversary can only access the outputs of DNNs. Hence attacking a model in the black-box setting is much more challenging. As recent studies have shown adversarial examples have a property of transferability, making it feasible to perform black-box attacks by using adversarial examples generated on a white-box model. Therefore, several efforts have been made to boost the transferability of adversarial examples \cite{dong2018boosting,xie2019improving}, in order to achieve high success rates of black-box attacks. Nevertheless, existing works focus on improving the transferability of image adversarial examples, while the transferability of video adversarial examples has not yet been explored. 

\begin{figure*}
    \centering
    \includegraphics[width=0.9\textwidth]{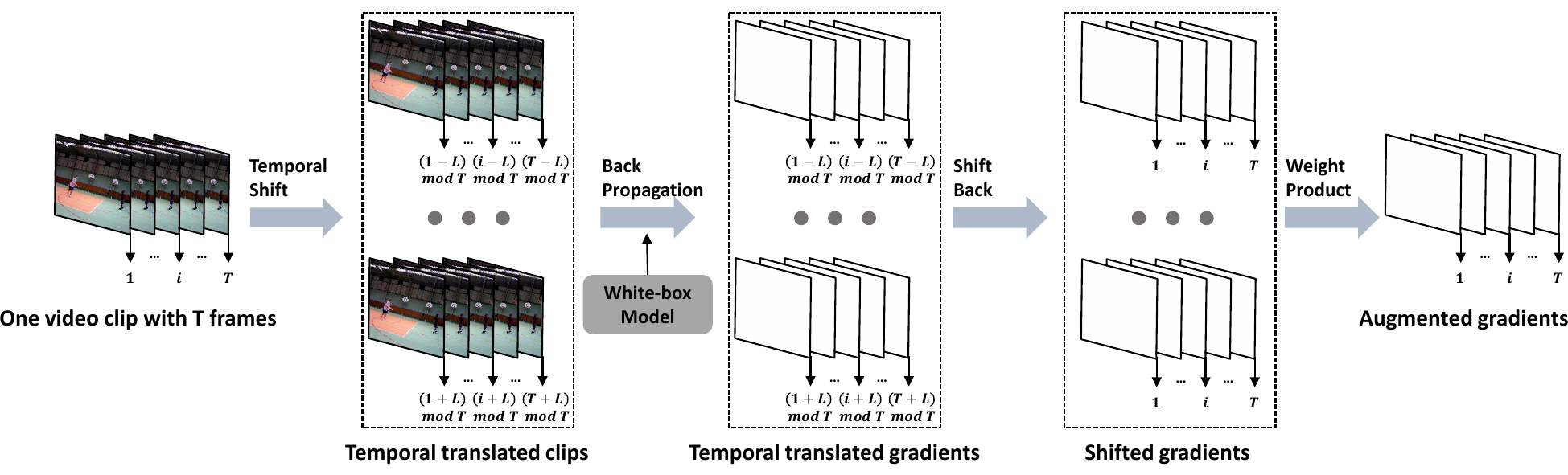}
    \caption{Overview of the proposed method. Given a video clip, temporal shifting is applied to generate a set of video clips. Then gradients obtained from these temporal translated video clips are linearly combined to form the augmented gradients. As temporal translation alleviates the effect of different discriminative temporal patterns between video recognition models, the adversarial video example generated from the augmented gradients is less sensitive to the discriminative temporal patterns of the white-box model and hence enjoys higher transferability.
    }
    \label{fig:overview}
\end{figure*}

This paper investigates the transferability of video adversarial examples to realize black-box attacks for video recognition models. The major challenge of this task comes from the fact that generated adversarial examples are prone to overfitting to white-box models and have poor transferability for other black-box models. Compared to images, videos have an additional temporal dimension, endowing videos with rich temporal information that describes the dynamic cues (\eg, motion). To capture such rich temporal information, various of video recognition models (\eg, I3D \cite{carreira2017quo}, SlowFast \cite{feichtenhofer2019slowfast}) which are based on 3D convolutional neural networks (CNNs) are designed. Intuitively, the discriminative temporal patterns captured by video recognition models may vary across different architectures. As generated adversarial examples are highly correlated with such patterns or gradients of the white-box model, directly utilizing adversarial video examples generated from the white-box model to attack other black-box video recognition models may lead to unsatisfactory results. 

To improve the transferability of video adversarial examples, this paper proposes a temporal translation method to generate adversarial examples that generalize well across different video models. The key idea is to optimize adversarial examples on a set of temporal translated video clips. In this way,  video adversarial examples will be less sensitive to discriminative temporal patterns of white-box models and enjoy higher transferability. Our method is motivated by \cite{dong2019evading}, where spatial translations are adopted on images to mitigate the effect of different discriminative regions between models and improve the transferability of image adversarial examples. 

Figure \ref{fig:overview} gives an overview of the proposed method. Given a video clip, temporal translation with a shift length $L$ is applied to generate a set of video clips. Note that the temporal translation can be performed both forward and backward along the temporal axis, hence we can obtain $2L$ translated video clips. Along with the original video clip, these $2L+1$ video clips are then inputted into the white-box model to obtain corresponding translated gradients, which are the gradients of the loss function with respect to video clips. To obtain the augmented gradients, the translated gradients are then shifted back to the original temporal order and combined linearly by a weight matrix. Finally, the resulting augmented gradients are used to generate adversarial examples that are able to generalize across different recognition models. 
We briefly summarize our primary contributions as follows:
\begin{itemize}
    \item We study transfer-based attacks in videos and propose a temporal translation attack method to boost the transferability of adversarial samples. To the best of our knowledge, this is the first work on transfer-based black-box attacks for video recognition models.
    \item We provide insightful analysis on the correlation of discriminative temporal patterns from different models and empirically prove that the discriminative temporal patterns among video models are different. Based on this observation, we combine the gradients from temporal translated videos to generate adversarial examples with higher transferability.
    \item We conduct empirical evaluations using six video recognition models trained with the Kinetics-400 dataset and UCF-101 dataset. Extensive experiments demonstrate that our proposed method helps to boost the transferability of video adversarial examples by a large margin.
\end{itemize}

\section{Related Work}
In this section, we first introduce white-box attacks, then review several transfer-based attacks on image models. We then review existing works on black-box attacks on video recognition models for showing the necessity of proposing transfer-based video attacks. 
We finally present several video recognition models with varying structures.
\subsection{Transfer-based attacks on images models}
\label{related:image}
The basic white-box attacks: Fast Gradient Sign Method (FGSM) \cite{goodfellow2014explaining}, Basic Iterative Method (BIM) \cite{kurakin2016adversarial1} are usually utilized for transfer-based attacks. FGSM takes a one-step update on the clean sample along with the sign of gradient towards maximizing the loss function. Basic Iterative Method (BIM) iteratively applies FGSM multiple times with a small step size. Although BIM can overfit the discriminative pattern well and generate more powerful adversarial examples than FGSM on the white-box model, the transferability of generated adversarial examples is worse than FGSM \cite{kurakin2016adversarial2}.
To further improve the transferability of adversarial examples, several approaches are proposed recently. One of the main causes for low transferability is that the generated adversarial examples tend to over-fit the white-box model. Hence to avoid over-fitting, several methods propose to increase the diversity of the input data with data augmentation. 
For example, Variance-Reduced attack (VR Attack) \cite{wu2018understanding} adds gaussian noise to the input for alleviating the shattering of gradients. Diversity Input attack (DI Attack) \cite{xie2019improving} applies random resizing and padding to the input at each iteration for creating diverse input patterns. Translation-invariant attack (TI Attack) \cite{dong2019evading} optimizes perturbations over an ensemble of translated images to generate more transferable adversarial examples against defense models. Scale-invariant method (SIM Attack) \cite{lin2019nesterov} optimizes the perturbations over the scale copies of the input.
Other methods attempt to stabilize the update direction or to obtain the vulnerable (high transfer-based attack success rate) gradients for optimization. For example. Momentum Iterative attack (MI Attack) \cite{dong2018boosting} integrates the momentum term into the iterative process for stabilizing update directions. Skip Gradient method (SGM Attack) \cite{wu2020skip} uses more gradients from the skip connections rather than residual modules for improving the transferability. Nesterov accelerated gradient is integrated into the FGSM to boost the transferability \cite{lin2019nesterov}. Besides, new loss functions are also proposed for optimizing the transferablility. For example, Attention-guided Transfer Attack (ATA) \cite{wu2020boosting} maximizes distance between clean images and their adversarial images in the attention map. In contrast, we discover different discriminative temporal patterns among video recognition models, which we utilize to overcome the over-fitting problem and improve the transferability of video adversarial examples. Different to TI, which adopts translation on the spatial dimension, our method performs translation on the temporal dimension and shifts gradients of translated inputs back to the original order before taking the weighted sum.
To the best of our knowledge, our method is the first attempt in investigating the transfer-based attacks in the video domain.

\subsection{Black-box attacks on video recognition models}
\label{related:black}
The black-box setting assumes that the adversary can only access the output of models, hence it is much more difficult than the adversarial attack under the white-box scenario. To realize black-box attack on video recognition models, Jiang \etal~\cite{jiang2019black} estimate the partition-based rectifications by the NES \cite{ilyas2018black} on partitions of tentative perturbations transferred from image models to generate adversarial examples with fewer queries.
To boost the attack efficiency and reduce the query numbers, Wei \etal~\cite{wei2020heuristic} propose a heuristic black-box attack method to generate sparse adversarial perturbations. Later, Zhang \etal~\cite{zhang2020motion} introduce a motion-excited sampler to correlate pixels in videos. 
Chen \etal~\cite{chen2021attacking} stealthily attacks video models with bullet-screen comments.
However, these methods still need a lot of queries to attack video models. Different from these works, we propose to realize black-box attacks on video recognition models by improving the transferability of video adversarial examples generated on white-box models. 

%----------------------------------------------------------------
\begin{figure*}[ht]
\centering
\subfigure[Zero padding]{
    \centering
    \includegraphics[width=0.3\textwidth]{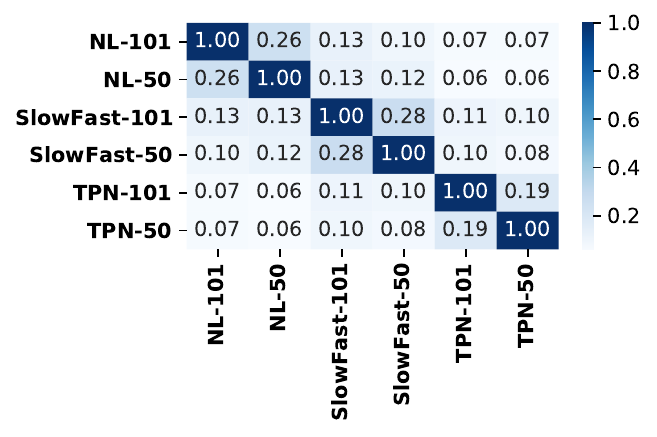}}
\subfigure[Mean padding]{
    \centering
    \includegraphics[width=0.3\textwidth]{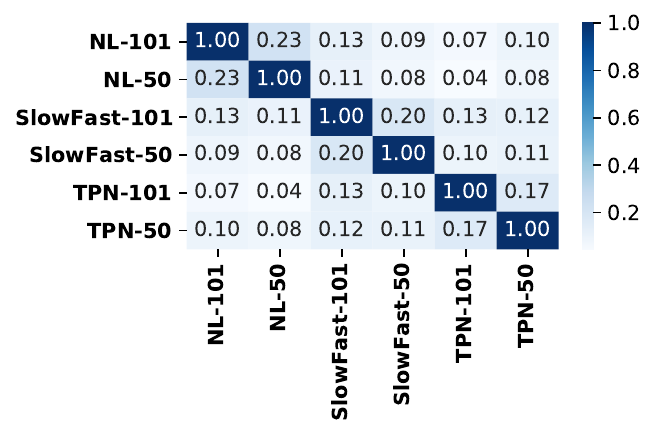}}
\subfigure[CAM]{
    \centering
    \includegraphics[width=0.3\textwidth]{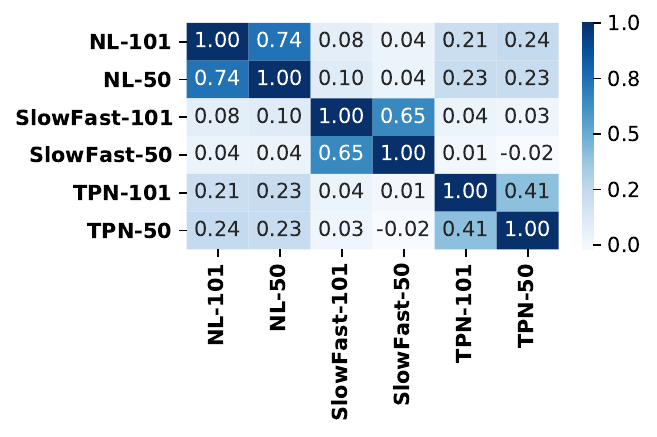}}
    
\caption{Correlations of discriminative temporal patterns between different video recognition models. A high correlation value indicates similar discriminative temporal patterns between the two models and is marked in the darker color.}
\label{fig:pattern}
\end{figure*}
%------------------------------------------------------------------------------------------

\subsection{Video recognition models}
\label{related:video}
Video recognition task has been widely investigated in recent years. Many studies \cite{jiang2017exploiting, yue2015beyond} applies 2D CNNs over per-frame input to extract features, followed by a 1D module that integrates temporal features. The separation of spatial and temporal modeling limits the ability to jointly capture the dynamic semantics of videos. Therefore, 3D CNNs are proposed to stack 3D convolutions to better handle temporal and spatial information. For example, I3D \cite{carreira2017quo} utilizes the inflated 2D convolution kernels to learn spatio-temporal representations from videos. Non-local network \cite{wang2018non} inserts the non-local blocks into I3D, non-local blocks calculate the response at a position as a weighted sum of all position features. SlowFast \cite{feichtenhofer2019slowfast} involves a slow pathway to capture spatial semantics, and a fast pathway to capture motion. SlowFast uses lateral connections to fuse from the fast to the slow pathway. This architecture contrasts the speed along the temporal axis. Temporal Pyramid Network (TPN) \cite{yang2020temporal} models the visual tempos of action instances as a feature-level pyramid. In the feature aggregation part, TPN fuses features from multiple depths in a single network to cover various visual tempos. There also exist other architectures of video recognition models, \eg, pseudo-3d residual network \cite{qiu2017learning}. In this paper, we use three representative video recognition models for experiments, including non-local network (NL), SlowFast, and TPN, with 3D ResNet-50 and ResNet-101 as backbones. For simplicity, we use Network-50/101 to denote the specified Network with ResNet50/ResNet101 as the backbone.

\section{Methodology}

\subsection{Preliminary}
Given a video sample $x \in \mathcal{X} \subset \mathbf{R}^{T \times H \times W \times C}$ with its ground-truth label $y \in \mathcal{Y} = \{1, 2, ..., K\}$, where $T$, $H$, $W$, $C$ denote the number of frames, height, width and channels respectively. K represents the number of classes. 
We use $f(x): \mathcal{X} \to \mathcal{Y} $ to denote the prediction for an input video. In this paper, we focus on untargeted adversarial attacks. Thus, the goal of adversarial attacks is to add an adversarial perturbation $\delta$ on $x$ to generate the adversarial example $x^{adv}$, which fools the DNNs to satisfy $f(x^{adv}) \neq y$. 
To make the perturbation human-imperceptible, the perturbation $\delta$ is restricted by $||\delta||_{p} \leq \epsilon$, where $||\cdot||_{p}$ denotes the $L_{p}$ norm, and $\epsilon$ is a constant of the norm constraint. In this paper, we set $L_{p}$ as $L_{\infty}$ norm. Denoting $J$ as the loss function, the objective of untargeted adversarial attacks can be formulated as follows: 
\begin{equation}
    \label{eq1}
        \mathop{\arg\max}_{\delta} J(f(x+\delta),y), \\
        s.t.\,\, ||\delta||_{\infty}<\epsilon.
\end{equation}
In white-box settings, the Equation \ref{eq1} can be approximately solved by an iterative attack process by computing the gradients of $J$ with respect to $x+\delta$. However, in black-box settings, the adversary cannot access the gradients. This paper focuses on transfer-based black-box attacks, which leverage adversarial examples generated from a white-box video recognition model to attack other black-box video recognition models.

\subsection{Discriminative temporal patterns analysis}
\label{ctp3.2}
Our proposed method is based on the assumption that different video recognition models rely on different discriminative temporal patterns, and hence adversarial videos generated from white-box models are difficult to transfer to other video recognition models. To verify this assumption and explain the reason for the low transferability of video adversarial examples, we analyze discriminative temporal patterns of different video recognition models in this section.

In contrast to image recognition models, where discriminative regions can be directly visualized with class activation mapping \cite{zhou2016learning}, discriminative temporal patterns of video recognition models are difficult to visualize. As a result, directly comparing the discriminative temporal patterns of different video models through visualization is not feasible. To this end, we compare discriminative temporal patterns in an indirect way. Intuitively, if two models share similar temporal patterns, the distribution of the importance of frames would be similar. Therefore, the discriminative temporal patterns can be compared by calculating the correlations between the importance orders of video frames from different models. In this work, we provide three ways to measure the importance of each frame: Grad-CAM \cite{selvaraju2017grad}, zero-padding, and Mean padding.

\textbf{Grad-CAM.} Grad-CAM uses the gradients of the predicted value in the target class with respect to the final convolutional layers to highlight the important regions. After obtaining the attention map by Grad-CAM (Please refer to \cite{selvaraju2017grad} for details), the average value of $i$-th frame's attention map is used to present the $i$-th frame's importance $p_i$.

\textbf{Zero-padding.} Zero-padding measures the contribution of $i$-th frame by replacing the $i$-th frame with zero values. Let $M_{i} \in \{0, 1\}^{T \times H \times W \times C}$ denote the temporal mask, where elements in $i$-th frame are 0, other frames are 1. Thus, the $i$-th frame's importance $p_{i}$ is defined as follows:
\begin{equation}
    \label{eq2}
    p_{i} = J(f(x \cdot M_{i}),y) - J(f(x),y).
\end{equation}
If the value of $p_{i}$ is large, it indicates the $i$-th frame is important for the model. 

\textbf{Mean-padding.} In contrast to zero-padding, mean-padding measures the contribution of $i$-th frame by replacing the $i$-th frame with the mean value of the previous and the next frames. Similar to zero-padding, the importance of $i$-th frame is calculated by measuring the loss change before and after mean-padding.

Then, we can obtain the model $A$'s importance list $P_{x}^{A} = \{p_{1}, p_{2}, ..., p_{T}\}$ for each frame in the video $x$. The correlation between any two models ($A$ and $B$) can be calculated by the Spearman's Rank Correlation \cite{zwillinger1999crc} between $P_{x}^{A}$ and $P_{x}^{B}$, which is defined as follows:
\begin{equation}
    \label{eq3}
     d_j = rg(P_x^A) - rg(P_x^B), \\
    \rho_{A,B} = 1-\frac{6\sum_{j=1}^{T}d_{j}^{2}}{T(T^{2}-1)}, 
\end{equation}
where the function $rg(\cdot)$ performs a sorting process and returns the importance orders of video frames. The value of $\rho_{A,B}$ varies between -1 and +1. $\rho_{A,B}$ equals 0 indicates there is no correlation between discriminative temporal patterns of model A and model B. $\rho_{A,B}$ will be large when $rg(p_x^A)$ and $rg(p_x^B)$ have a similar rank and be small when $rg(p_x^A)$ and $rg(p_x^B)$ have a dissimilar rank.
$\rho_{A,B}$ equals -1 or +1 imply an exact monotonic relationship between $rg(p_x^A)$ and $rg(p_x^B)$. In this way, the correlation of discriminative temporal patterns can be measured by $\rho_{A,B}$.

We analyze the correlations of discriminative temporal patterns among 6 video recognition models, including NL-101, NL-50, SlowFast-101, SlowFast-50, TPN-101, and TPN-50.
This analysis is conducted on 400 randomly selected video clips from the Kinetics-400 dataset.
Figure \ref{fig:pattern} shows their average correlations of the discriminative temporal patterns. It can be clearly observed that the correlations of discriminative temporal patterns between different models are low, which verifies our assumption that there are different temporal patterns among video recognition models. Therefore, disturbing the temporal discriminative pattern of the white-box model being attacked can help to improve the transferability of video adversarial examples.

\subsection{Temporal translation attack method}
\label{ctp3.3}
To transfer adversarial examples across video models, we propose the temporal translation attack method based on the previous observation. Through translating video clips along the temporal axis, we can obtain multiple gradients from various temporal directions. It helps to alleviate the high correlation between generated adversarial examples and the white-box model being attacked.

Therefore, in our approach, instead of using $\nabla_{x}J(f(x+\delta), y)$ as gradients to iteratively update adversarial examples, we combine gradients of temporally translated videos by $\bf{g}$:
\begin{equation}
    \label{eq4}
    {\bf g} = \sum_{i=-L}^{L}\omega_{i}\mathit{TR}_{-i}(\nabla_{x}J(f(\mathit{TR}_{i}(x+\delta)), y)),
\end{equation}
where $L$ denotes the shift length, which is the maximal shift number of frames, and $i \in \{-L, ..., L\}$. The temporal translation function $\mathit{TR}_{i}(\cdot)$ shifts each video frame by $i$ frames along the temporal axis, which generates temporally translated video clips and is carried out in the video loop. We then shift the computed gradients back $\mathit{TR}_{-i}(\nabla_{x}J(f(\mathit{TR}_{i}(x+\delta)), y))$ with the inverse function, converting sequences of temporally translated gradients $\nabla_{x}J(f(\mathit{TR}_{i}(x+\delta)), y)$ back to the same sequence as $x$. $W=\{\omega_{-L}, ..., \omega_{0}, ..., \omega_{L}\}$ is the symmetric weight matrix of size $2L+1$, with $\omega_{i} = \omega_{-i}$. $\omega_{i} \in W$ is the weight for $\mathit{TR}_{-i}(\nabla_{x}J(f(\mathit{TR}_{i}(x+\delta)), y))$.
The Equation \ref{eq4} calculates gradients for all translated clips and then combines these gradients with the weight matrix $W$.
In this way, the temporal translation function makes it possible to optimize an adversarial example using a set of temporally translated video clips with varying positions, which prevents over-fitting. 
With this method, the generated video adversarial example has a low correlation with the discriminative temporal patterns used by the white-box model and can be better transferred to another model. 
The fixed weight matrix $W$ determines the quality of augmented gradients. Intuitively, the video clips with smaller shifts should be more important than those with larger shifts. Following~\cite{dong2019evading}, we consider three strategies to generate $W$: 
\begin{itemize}
    \item [1)] Uniform: $\omega_{i} = \frac{1}{2L+1}$;
    \item [2)] Linear: $\omega_{i} = 1-\frac{|i|}{2L+1}$, and $\omega_{i} = \frac{\omega_{i}}{\sum_{i}\omega_{i}}$;
    \item [3)] Gaussian: $\omega_{i} = \frac{1}{\sqrt{2\pi}\sigma}e^{-\frac{i^{2}}{2\sigma^{2}}}$, and $\omega_{i} = \frac{\omega_{i}}{\sum_{i}\omega_{i}}$. Because the radius of $3\sigma$ has more than 97\% of the total information, we assign the standard deviation $\sigma$ to $\frac{L}{3}$.
\end{itemize}

It is worthwhile to note that for temporal translation, apart from adjacent shifting which shifts the frames within the adjacent positions, other frame shifting strategies, such as random shifting, remote shifting, can also be applied. Random shifting shifts frames to random positions while remote shifting shifts the clip by $i +\frac{T}{2}$ frames along the temporal axis. In our approach, we choose adjacent shifting in order to avoid a great change of gradients caused by large shifts. We will discuss the effect of different shifting strategies in Section \ref{exp:ablation}. 

Algorithm \ref{alg} illustrates the generation of adversarial examples. In each update step, we combine the translated gradients by the weight matrix $W$, then use $clip_{x,\epsilon}(\cdot)$ to restrict generated adversarial examples to be within the $\epsilon$-ball of x. 
The algorithm introduces the temporal translation-based gradient augmentation to the procedure of FGSM ($I=1$) and BIM ($I>1$), named TT-FGSM and TT-BIM separately. Our algorithm focuses on the temporal dimension, and thus it can be easily integrated into other spatio-based transfer methods.

\begin{algorithm}[tb]
\caption{Temporal translation (TT) attack}
\label{alg}
\textbf{Input}: The loss function $J$, clean video $x$, ground-truth class $y$.\\
\textbf{Parameter}: The perturbation budget $\epsilon$, iteration number I, shift $L$, weight matrix $W$.\\
\textbf{Output}: The adversarial example.
\begin{algorithmic}[1] %[1] enables line numbers
\STATE $x_{0} \gets x$
\STATE $\alpha \gets \frac{\epsilon}{I}$
\FOR{$i = 0$ to $I-1$}
\STATE $x_{i+1} = clip_{x,\epsilon}(x_{i} + \alpha \cdot {\bf g} )$
\ENDFOR

\STATE \textbf{return} $x_{I}$ 
\end{algorithmic}
\end{algorithm}

\section{Experiments}

\subsection{Experimental setting}
\label{exp:set}
\textbf{Dataset.} 
We evaluate our approach using UCF-101 \cite{soomro2012ucf101} and Kinetics-400 datasets \cite{kay2017kinetics}, which are widely used datasets for video recognition. 
UCF-101 consists of 13,320 videos from 101 action categories. 
Kinetics-400 contains approximately 240,000 videos from 400 human action classes.

\textbf{Video recognition models.}
Our proposed method is evaluated on three video recognition models\footnote{The models are implemented in \url{https://cv.gluon.ai/model_zoo/action_recognition.html}}: Non-local network (NL) \cite{wang2018non}, SlowFast \cite{feichtenhofer2019slowfast} and TPN \cite{yang2020temporal}. For each model, we experiment with ResNet-50 and ResNet-101 as backbones, totaling 6 models. Non-local network inserts 5 non-local blocks into I3D. All models are trained on the RGB domain. Input clips are formed by randomly cropping out 64 consecutive frames from videos and then skipping every other frame. The spatial size of the input is $224\times224$.

\textbf{Attack setting.}
In our experiments, video recognition models with ResNet101 as its backbone are used as white-box models for adversarial example generation. We use the Attack Success Rate (ASR) to evaluate the attack performance, which is the rate of adversarial examples that are successfully misclassified by the black-box video recognition model. Thus higher ASR means better adversarial transferability. We randomly sampled one clip, which is correctly classified by all models, for each class from the kinetics-400 validation dataset and the UCF-101 test dataset to conduct all attacks. Following \cite{dong2019evading,zhou2018transferable}, we set the maximum perturbation as $\epsilon=16$ for all experiments. For the iterative attack, we set the iteration number to $I=10$, and thus the step size $\alpha = 1.6$. 

\subsection{Performance Comparison}
%--------------------------------------------------------------------
\label{exp:comparison}

\begin{figure*}[h]
\centering
    \subfigure[UCF-101]{
        % \centering
        \includegraphics[width=\textwidth]{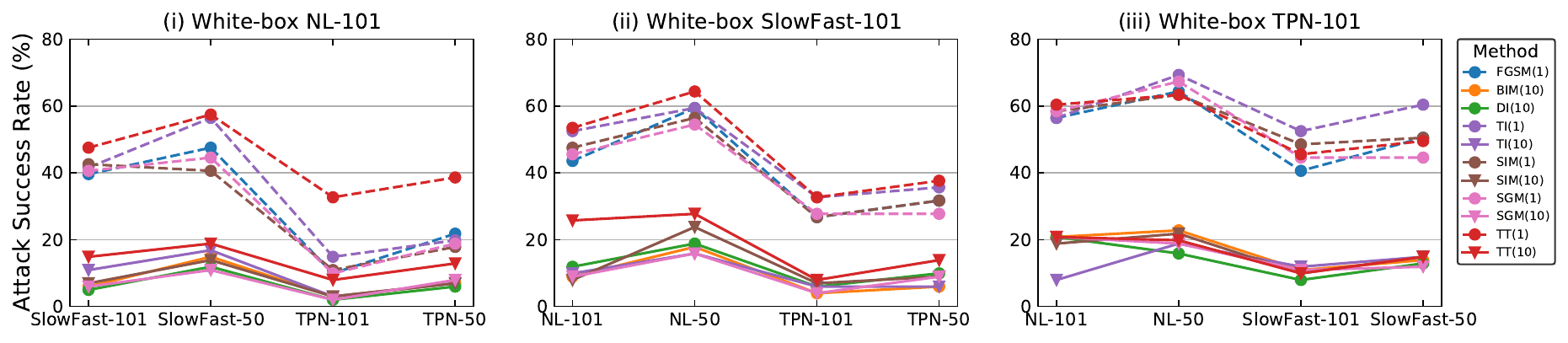}
        \label{com_ucf}
    }
    \subfigure[Kinetics-400]{
        % \centering
        \includegraphics[width=\textwidth]{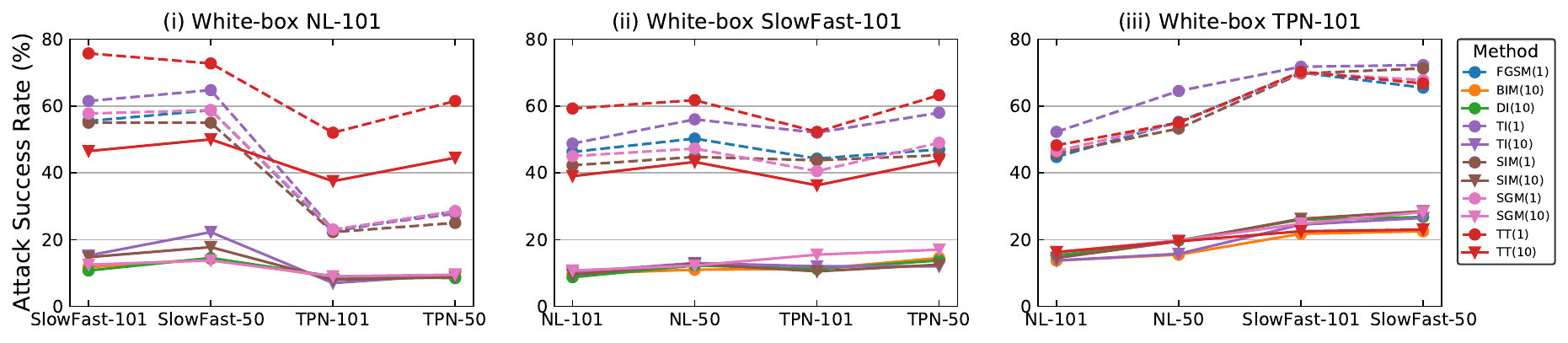}
        \label{com_kinetics}
    }
\caption{Attack success rates (\%) against video recognition models. The top and bottom row are the results on the UCF-101 and Kinetics-400 respectively. The three columns use the NL-101, SlowFast-101, and TPN-101 models as white-box models separately. Dashed lines and solid lines denote the one-step and multi-step attack methods respectively. The number in parentheses indicates the number of iterations. In the legend, our methods are in red color.}
\label{fig_per_com}
\end{figure*}

\begin{table}[]
    \centering
    \begin{tabular}{l c c} \toprule
         Method & UCF-101 & Kinetics \\ \midrule
         TI(1) & 45.96 & 54.38\\ 
         TI+TT(1) & \textbf{52.48} & \textbf{63.22}\\  \midrule
         ATA(10) & 9.98 & 8.52 \\ 
         ATA+TT(10) & \textbf{12.87} & \textbf{10.58}\\  \midrule
         MI(10) & 33.58 & 40.60\\ 
         MI+TT(10) & \textbf{44.80} & \textbf{63.50}\\ \bottomrule 
         
    \end{tabular}
    \caption{The average ASR of all black-box video models under three attack settings using NL-101, SlowFast-101 and TPN-101 as white box models, respectively. Best results are in bold.
    }
    \label{tab:1}
\end{table}

In this section, we first compare our method with several baselines such as FGSM and BIM and other methods that aims to avoid over-fitting to the white-box model, including DI, TI, SIM, and SGM. Then we integrate our method with other attack methods, such as TI, MI, and ATA, for showing the compatibility of our method. In the experiments, we follow the original settings of these baseline methods. NL-101, SlowFast-101 and TPN-101 are used as white-box models for generating adversarial examples to attack other black-box models. For our method, the shift length $L$ is set as 7, the weight matrix $W$ is generated with Gaussian function, and the adjacent shift is adopted in the temporal translation. We conduct performance comparison under both one-step attack and iterative attack on UCF-101 and Kinetics-400. The results are summarized in Figure~\ref{fig_per_com}, where dashed lines and solid lines denote the one-step attaks and multi-step attacks respectively, the number in parentheses of the legend indicates the number of iterations.
We use TT(1) and TT(10) to denote TT-FGSM and TT-BIM, respectively.

\textbf{One-step attack.} 
For the one-step attack, we have the following observations. 
First, except for using TPN-101 as the white-box model (Figure 3(i), 3(ii)), our temporal translation attack method (TT(1)) consistently achieves much higher success rates than other baselines. For example, when attacking TPN-101 and TPN-50 using adversarial examples generated from NL-101, our method yields an improvement of more than 30\% compared with other transfer-based attack methods. The results basically demonstrate the effectiveness of our method towards improving the transferability of video adversarial examples. Second, when using TPN-101 as the white-box model for adversarial example generation (Figure 3(iii)), the improvement of attack success rate gained from temporal translation is not so significant. 
Compared to FGSM(1), the performance improvement is less than 1\% when attacking SlowFast-101. The results basically suggest TPN-101 is less sensitive to temporal patterns.

\textbf{Iterative attack.}
For the iterative attack, similar trends can be observed. First, when using NL-101 (Figure 3(i)) and SlowFast-101 (Figure 3(ii)) as white-box models, our method (TT(10)) outperforms all the other baselines. Compared to BIM(10), our temporal translation boosts the attack success rates of more than 30\% for all the black-box models. The results verify the effectiveness of temporal translation in boosting the transferability of video adversarial examples. 
Second, similar to the results on the one-step attack, temporal translation (TT(10)) does not improve the success rate too much when using TPN-101 (Figure 3(ii)) as the white-box model.
Note that compared to adversarial examples generated with the one-step attack, adversarial examples generated with the iterative attack attain much lower success rates in black-box attacks when attacking video models. However iterative attacks (such as MI, DI, SIM, etc.) transfer better than one-step attacks when attacking image models. It indicates that iterative attacks of spatial domain can't directly be applied to temporal domain and iterative attacks tend to overfit white-box models in the high-dimensional video domain.
Nevertheless, combining with MI, the iterative attack MI+TT(10) can exceed the one-step attack.

\textbf{Combining with existing methods.}
Figure~\ref{fig_per_com}(iii) shows that TI(1) achieves better results than TT, this is because TPN-101 is less sensitive to temporal patterns. However, TT can also play a supplementary role to TI. For verify this, we evaluate the performance of TI-TT(1), which combines the temporal translation with translation-invariant (TI) in the one-step attack. In addition, we also evaluate the performance of two variants of our method, denoted as TT-MI(10), TT-ATA(10) for proving that our method is compatible with other attacks. 
From the results shown in Table \ref{tab:1}, we observe that by combining temporal translation with other works, our methods achieve the best results. It demonstrates that temporal translation plays a supplementary role to other transfer-based attack methods, in terms of improving the transferability of video adversarial examples. In addition, TI+TT(1) achieves the best result by mitigating the spatial and temporal discriminative patterns in the one-step attack. In the iterative attack, MI+TT(10) exceeds the results of the one-step attack on Kinetics-400 through the integration of momentum and temporal translation. Moreover, ATA(10) achieves the worst result. This is because different discriminative temporal patterns induce various attention maps in video models, while ATA(10) depends on similar attention maps.
To summarize, the results demonstrate our temporal translation method is effective in increasing the success rates of transfer-based attack, posing threatens to the real-world applications of video recognition models.

\begin{figure}
    \centering
    \includegraphics[width=0.7\columnwidth]{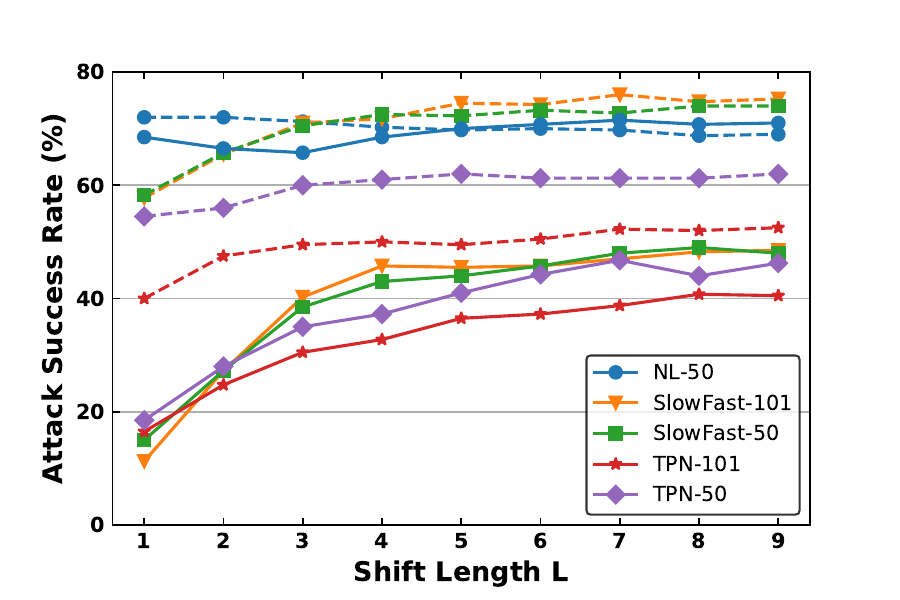}
    \caption{Attack success rates of temporal translation based attacks with various shift length $L$. Dashed lines and solid lines denote TT(1) and TT(10) respectively.}
    \label{fig:kernlen}
\end{figure}

\begin{table}[]
    \centering
    \begin{tabular}{l c c c} \toprule
        Components of TT & Choice &  TT(1) & TT(10)  \\ \midrule
        \multirow{3}{*}{Weight matrix} & Uniform &  65.06 & 49.80\\ 
         & Linear & 66.37 & \textbf{52.60}\\ 
         & Gaussian & \textbf{66.80} &51.35\\ \midrule
        \multirow{3}{*}{Shifting strategies} & Adjacent & \textbf{65.56} & \textbf{43.37} \\ 
         & Random & 58.06 & 23.25\\ 
         & Remote & 57.31 &26.81 \\ \bottomrule
         
    \end{tabular}
    \caption{Average attack success rate for TT(1) and TT(10) with different weight matrices and shifting strategies. Best results are in bold.}
    \label{tab:2}
\end{table}

\subsection{Ablation study}
\label{exp:ablation}
We investigate the effects of different shift lengths $L$, weight matrices $W$ and shifting strategies on temporal translation. 
The evaluations are conducted with TT(1) and TT(10).
We use NL-101 model as the white-box model to evaluate the performances.

\textbf{Shift length.} 
The shift length $L$ determines how many translated clips are used for temporal gradient augmentation. When $L=0$, the attacks based on temporal translation degenerate into their vanilla versions. Thus, it is vital to study $L$. Figure \ref{fig:kernlen} shows the results, where $L$ ranges from 1 to 9, dashed lines and solid lines denote TT(1) and TT(10) respectively. As can be seen, the attack success rates on NL-50 are rather high and remain stable under different values of $L$ for both TT(1) and TT(10). In contrast, for other models, the success rates increase firstly and then tend to be stable when increasing the value of $L$.
This is because NL-50 shares the similar architecture with NL-101.
Although a large value of $L$ leads to better transferability, it requires a large computation cost. To balance attack success rates and the computational cost, we set $L$ as 7 in our experiments.

\textbf{Weight matrix.} We set the shift length $L$ as 7 for comparing the performance of different weight matrices: uniform, linear and gaussian. Table \ref{tab:2} shows the results. It can be seen that for TT(1), the gaussian matrix outperforms other matrices on average ASR. And for TT-BIM, the linear matrix achieves better results than the uniform and gaussian matrices. In general, the linear and the gaussian matrices outperform the uniform matrix.
Thus, we empirically adopt the gaussian matrix because of similar results between the linear and the gaussian matrices.

\textbf{Shifting strategies.}
Table \ref{tab:2} shows the results of TT(1) and TT(10) with different shifting strategies: adjacent shifting, random shifting, and remote shifting. The adjacent shifting leads to higher attack success rates than both random and remote shiftings. This is because adjacent shifting provides a more stable gradient for optimization compared to random and remote shiftings.

\section{Conclusion}
In this paper, we propose a temporal translation attack method, which generates adversarial examples by using a set of temporal translated videos to mitigate the effect of different discriminative temporal patterns across video recognition models. Our method can be directly combined with other transfer-based attack methods and the generated adversarial examples have higher transferability against video recognition models. Extensive experimental results have demonstrated the effectiveness of our method and indicated the vulnerability of video recognition models in the transfer-based attack setting. Since our attack generates perturbations without temporal consistent, one of the potential defense methods can be designed by checking the temporal consistent. We will explore these in future works.

\section{Acknowledgments}
The authors would like to thank the anonymous referees for their valuable comments and helpful suggestions. This work was supported in part by NSFC project (\#62032006), Science and Technology Commission of Shanghai Municipality Project (20511101000), and in part by Shanghai Pujiang Program (20PJ1401900).

\bibliography{refer.bib}

\begin{thebibliography}{42}
\providecommand{\natexlab}[1]{#1}

\bibitem[{Carreira and Zisserman(2017)}]{carreira2017quo}
Carreira, J.; and Zisserman, A. 2017.
\newblock Quo vadis, action recognition? a new model and the kinetics dataset.
\newblock In \emph{proceedings of the IEEE Conference on Computer Vision and
  Pattern Recognition}, 6299--6308.

\bibitem[{Chen et~al.(2020{\natexlab{a}})Chen, Pan, Wei, Wang, Ngo, and
  Chua}]{chen2020zero}
Chen, J.; Pan, L.; Wei, Z.; Wang, X.; Ngo, C.-W.; and Chua, T.-S.
  2020{\natexlab{a}}.
\newblock Zero-shot ingredient recognition by multi-relational graph
  convolutional network.
\newblock In \emph{Proceedings of the AAAI Conference on Artificial
  Intelligence}, volume~34, 10542--10550.

\bibitem[{Chen et~al.(2020{\natexlab{b}})Chen, Zhu, Ngo, Chua, and
  Jiang}]{chen2020study}
Chen, J.; Zhu, B.; Ngo, C.-W.; Chua, T.-S.; and Jiang, Y.-G.
  2020{\natexlab{b}}.
\newblock A Study of Multi-Task and Region-Wise Deep Learning for Food
  Ingredient Recognition.
\newblock \emph{IEEE Transactions on Image Processing}, 30: 1514--1526.

\bibitem[{Chen et~al.(2021)Chen, Wei, Chen, Wu, and Jiang}]{chen2021attacking}
Chen, K.; Wei, Z.; Chen, J.; Wu, Z.; and Jiang, Y.-G. 2021.
\newblock Attacking Video Recognition Models with Bullet-Screen Comments.
\newblock \emph{arXiv preprint arXiv:2110.15629}.

\bibitem[{Dong et~al.(2018)Dong, Liao, Pang, Su, Zhu, Hu, and
  Li}]{dong2018boosting}
Dong, Y.; Liao, F.; Pang, T.; Su, H.; Zhu, J.; Hu, X.; and Li, J. 2018.
\newblock Boosting adversarial attacks with momentum.
\newblock In \emph{Proceedings of the IEEE conference on computer vision and
  pattern recognition}, 9185--9193.

\bibitem[{Dong et~al.(2019)Dong, Pang, Su, and Zhu}]{dong2019evading}
Dong, Y.; Pang, T.; Su, H.; and Zhu, J. 2019.
\newblock Evading defenses to transferable adversarial examples by
  translation-invariant attacks.
\newblock In \emph{Proceedings of the IEEE/CVF Conference on Computer Vision
  and Pattern Recognition}, 4312--4321.

\bibitem[{Feichtenhofer et~al.(2019)Feichtenhofer, Fan, Malik, and
  He}]{feichtenhofer2019slowfast}
Feichtenhofer, C.; Fan, H.; Malik, J.; and He, K. 2019.
\newblock Slowfast networks for video recognition.
\newblock In \emph{Proceedings of the IEEE/CVF International Conference on
  Computer Vision}, 6202--6211.

\bibitem[{Feng et~al.(2021)Feng, Zhang, He, Zhang, and
  Chua}]{feng2021empowering}
Feng, F.; Zhang, J.; He, X.; Zhang, H.; and Chua, T.-S. 2021.
\newblock Empowering Language Understanding with Counterfactual Reasoning.
\newblock \emph{arXiv preprint arXiv:2106.03046}.

\bibitem[{Goodfellow, Shlens, and Szegedy(2014)}]{goodfellow2014explaining}
Goodfellow, I.~J.; Shlens, J.; and Szegedy, C. 2014.
\newblock Explaining and harnessing adversarial examples.
\newblock \emph{arXiv preprint arXiv:1412.6572}.

\bibitem[{Guo et~al.(2021)Guo, Wei, Wang, and Zhang}]{guo2021meaningful}
Guo, Y.; Wei, X.; Wang, G.; and Zhang, B. 2021.
\newblock Meaningful Adversarial Stickers for Face Recognition in Physical
  World.
\newblock \emph{arXiv preprint arXiv:2104.06728}.

\bibitem[{He et~al.(2016)He, Zhang, Ren, and Sun}]{he2016deep}
He, K.; Zhang, X.; Ren, S.; and Sun, J. 2016.
\newblock Deep residual learning for image recognition.
\newblock In \emph{Proceedings of the IEEE conference on computer vision and
  pattern recognition}, 770--778.

\bibitem[{Ilyas et~al.(2018)Ilyas, Engstrom, Athalye, and Lin}]{ilyas2018black}
Ilyas, A.; Engstrom, L.; Athalye, A.; and Lin, J. 2018.
\newblock Black-box adversarial attacks with limited queries and information.
\newblock In \emph{International Conference on Machine Learning}, 2137--2146.
  PMLR.

\bibitem[{Jiang et~al.(2019)Jiang, Ma, Chen, Bailey, and
  Jiang}]{jiang2019black}
Jiang, L.; Ma, X.; Chen, S.; Bailey, J.; and Jiang, Y.-G. 2019.
\newblock Black-box adversarial attacks on video recognition models.
\newblock In \emph{Proceedings of the 27th ACM International Conference on
  Multimedia}, 864--872.

\bibitem[{Jiang et~al.(2017)Jiang, Wu, Wang, Xue, and
  Chang}]{jiang2017exploiting}
Jiang, Y.-G.; Wu, Z.; Wang, J.; Xue, X.; and Chang, S.-F. 2017.
\newblock Exploiting feature and class relationships in video categorization
  with regularized deep neural networks.
\newblock \emph{IEEE transactions on pattern analysis and machine
  intelligence}, 40(2): 352--364.

\bibitem[{Jiao et~al.(2021)Jiao, Jie, Luo, Chen, Jiang, Wei, and
  Ma}]{jiao2021two}
Jiao, Y.; Jie, Z.; Luo, W.; Chen, J.; Jiang, Y.-G.; Wei, X.; and Ma, L. 2021.
\newblock Two-stage Visual Cues Enhancement Network for Referring Image
  Segmentation.
\newblock In \emph{Proceedings of the 29th ACM International Conference on
  Multimedia}, 1331--1340.

\bibitem[{Kay et~al.(2017)Kay, Carreira, Simonyan, Zhang, Hillier,
  Vijayanarasimhan, Viola, Green, Back, Natsev et~al.}]{kay2017kinetics}
Kay, W.; Carreira, J.; Simonyan, K.; Zhang, B.; Hillier, C.; Vijayanarasimhan,
  S.; Viola, F.; Green, T.; Back, T.; Natsev, P.; et~al. 2017.
\newblock The kinetics human action video dataset.
\newblock \emph{arXiv preprint arXiv:1705.06950}.

\bibitem[{Kurakin, Goodfellow, and Bengio(2016)}]{kurakin2016adversarial2}
Kurakin, A.; Goodfellow, I.; and Bengio, S. 2016.
\newblock Adversarial machine learning at scale.
\newblock \emph{arXiv preprint arXiv:1611.01236}.

\bibitem[{Kurakin et~al.(2016)Kurakin, Goodfellow, Bengio
  et~al.}]{kurakin2016adversarial1}
Kurakin, A.; Goodfellow, I.; Bengio, S.; et~al. 2016.
\newblock Adversarial examples in the physical world.

\bibitem[{Lin et~al.(2019)Lin, Song, He, Wang, and Hopcroft}]{lin2019nesterov}
Lin, J.; Song, C.; He, K.; Wang, L.; and Hopcroft, J.~E. 2019.
\newblock Nesterov accelerated gradient and scale invariance for adversarial
  attacks.
\newblock \emph{arXiv preprint arXiv:1908.06281}.

\bibitem[{Liu et~al.(2020)Liu, Chen, Pan, Ngo, Chua, and
  Jiang}]{liu2020hyperbolic}
Liu, S.; Chen, J.; Pan, L.; Ngo, C.-W.; Chua, T.-S.; and Jiang, Y.-G. 2020.
\newblock Hyperbolic visual embedding learning for zero-shot recognition.
\newblock In \emph{Proceedings of the IEEE/CVF Conference on Computer Vision
  and Pattern Recognition}, 9273--9281.

\bibitem[{Liu et~al.(2018)Liu, Xia, Liu, He, Zhang, and
  Zimmermann}]{liu2018toward}
Liu, Z.; Xia, Y.; Liu, Q.; He, Q.; Zhang, C.; and Zimmermann, R. 2018.
\newblock Toward personalized activity level prediction in community question
  answering websites.
\newblock \emph{ACM Transactions on Multimedia Computing, Communications, and
  Applications (TOMM)}, 14(2s): 1--15.

\bibitem[{Qiu, Yao, and Mei(2017)}]{qiu2017learning}
Qiu, Z.; Yao, T.; and Mei, T. 2017.
\newblock Learning spatio-temporal representation with pseudo-3d residual
  networks.
\newblock In \emph{proceedings of the IEEE International Conference on Computer
  Vision}, 5533--5541.

\bibitem[{Ren et~al.(2016)Ren, He, Girshick, and Sun}]{ren2016faster}
Ren, S.; He, K.; Girshick, R.; and Sun, J. 2016.
\newblock Faster R-CNN: towards real-time object detection with region proposal
  networks.
\newblock \emph{IEEE transactions on pattern analysis and machine
  intelligence}, 39(6): 1137--1149.

\bibitem[{Selvaraju et~al.(2017)Selvaraju, Cogswell, Das, Vedantam, Parikh, and
  Batra}]{selvaraju2017grad}
Selvaraju, R.~R.; Cogswell, M.; Das, A.; Vedantam, R.; Parikh, D.; and Batra,
  D. 2017.
\newblock Grad-cam: Visual explanations from deep networks via gradient-based
  localization.
\newblock In \emph{Proceedings of the IEEE international conference on computer
  vision}, 618--626.

\bibitem[{Song et~al.(2021)Song, Chen, Wu, and Jiang}]{song2021spatial}
Song, X.; Chen, J.; Wu, Z.; and Jiang, Y.-G. 2021.
\newblock Spatial-temporal Graphs for Cross-modal Text2Video Retrieval.
\newblock \emph{IEEE Transactions on Multimedia}.

\bibitem[{Soomro, Zamir, and Shah(2012)}]{soomro2012ucf101}
Soomro, K.; Zamir, A.~R.; and Shah, M. 2012.
\newblock UCF101: A dataset of 101 human actions classes from videos in the
  wild.
\newblock \emph{arXiv preprint arXiv:1212.0402}.

\bibitem[{Szegedy et~al.(2013)Szegedy, Zaremba, Sutskever, Bruna, Erhan,
  Goodfellow, and Fergus}]{szegedy2013intriguing}
Szegedy, C.; Zaremba, W.; Sutskever, I.; Bruna, J.; Erhan, D.; Goodfellow, I.;
  and Fergus, R. 2013.
\newblock Intriguing properties of neural networks.
\newblock \emph{arXiv preprint arXiv:1312.6199}.

\bibitem[{Tian et~al.(2018)Tian, Pei, Jana, and Ray}]{tian2018deeptest}
Tian, Y.; Pei, K.; Jana, S.; and Ray, B. 2018.
\newblock Deeptest: Automated testing of deep-neural-network-driven autonomous
  cars.
\newblock In \emph{Proceedings of the 40th international conference on software
  engineering}, 303--314.

\bibitem[{Wang et~al.(2018)Wang, Girshick, Gupta, and He}]{wang2018non}
Wang, X.; Girshick, R.; Gupta, A.; and He, K. 2018.
\newblock Non-local neural networks.
\newblock In \emph{Proceedings of the IEEE conference on computer vision and
  pattern recognition}, 7794--7803.

\bibitem[{Wei et~al.(2019)Wei, Zhu, Yuan, and Su}]{wei2019sparse}
Wei, X.; Zhu, J.; Yuan, S.; and Su, H. 2019.
\newblock Sparse adversarial perturbations for videos.
\newblock In \emph{Proceedings of the AAAI Conference on Artificial
  Intelligence}, volume~33, 8973--8980.

\bibitem[{Wei et~al.(2020)Wei, Chen, Wei, Jiang, Chua, Zhou, and
  Jiang}]{wei2020heuristic}
Wei, Z.; Chen, J.; Wei, X.; Jiang, L.; Chua, T.-S.; Zhou, F.; and Jiang, Y.-G.
  2020.
\newblock Heuristic Black-Box Adversarial Attacks on Video Recognition Models.
\newblock In \emph{AAAI}, 12338--12345.

\bibitem[{Wu et~al.(2020{\natexlab{a}})Wu, Wang, Xia, Bailey, and
  Ma}]{wu2020skip}
Wu, D.; Wang, Y.; Xia, S.-T.; Bailey, J.; and Ma, X. 2020{\natexlab{a}}.
\newblock Skip connections matter: On the transferability of adversarial
  examples generated with resnets.
\newblock \emph{arXiv preprint arXiv:2002.05990}.

\bibitem[{Wu et~al.(2018)Wu, Zhu, Tai et~al.}]{wu2018understanding}
Wu, L.; Zhu, Z.; Tai, C.; et~al. 2018.
\newblock Understanding and enhancing the transferability of adversarial
  examples.
\newblock \emph{arXiv preprint arXiv:1802.09707}.

\bibitem[{Wu et~al.(2020{\natexlab{b}})Wu, Su, Chen, Zhao, King, Lyu, and
  Tai}]{wu2020boosting}
Wu, W.; Su, Y.; Chen, X.; Zhao, S.; King, I.; Lyu, M.~R.; and Tai, Y.-W.
  2020{\natexlab{b}}.
\newblock Boosting the transferability of adversarial samples via attention.
\newblock In \emph{Proceedings of the IEEE/CVF Conference on Computer Vision
  and Pattern Recognition}, 1161--1170.

\bibitem[{Wu et~al.(2020{\natexlab{c}})Wu, Li, Xiong, Jiang, and
  Davis}]{wu2020dynamic}
Wu, Z.; Li, H.; Xiong, C.; Jiang, Y.-G.; and Davis, L.~S. 2020{\natexlab{c}}.
\newblock A dynamic frame selection framework for fast video recognition.
\newblock \emph{IEEE Transactions on Pattern Analysis and Machine
  Intelligence}.

\bibitem[{Xie et~al.(2019)Xie, Zhang, Zhou, Bai, Wang, Ren, and
  Yuille}]{xie2019improving}
Xie, C.; Zhang, Z.; Zhou, Y.; Bai, S.; Wang, J.; Ren, Z.; and Yuille, A.~L.
  2019.
\newblock Improving transferability of adversarial examples with input
  diversity.
\newblock In \emph{Proceedings of the IEEE/CVF Conference on Computer Vision
  and Pattern Recognition}, 2730--2739.

\bibitem[{Yang et~al.(2020)Yang, Xu, Shi, Dai, and Zhou}]{yang2020temporal}
Yang, C.; Xu, Y.; Shi, J.; Dai, B.; and Zhou, B. 2020.
\newblock Temporal pyramid network for action recognition.
\newblock In \emph{Proceedings of the IEEE/CVF Conference on Computer Vision
  and Pattern Recognition}, 591--600.

\bibitem[{Yue-Hei~Ng et~al.(2015)Yue-Hei~Ng, Hausknecht, Vijayanarasimhan,
  Vinyals, Monga, and Toderici}]{yue2015beyond}
Yue-Hei~Ng, J.; Hausknecht, M.; Vijayanarasimhan, S.; Vinyals, O.; Monga, R.;
  and Toderici, G. 2015.
\newblock Beyond short snippets: Deep networks for video classification.
\newblock In \emph{Proceedings of the IEEE conference on computer vision and
  pattern recognition}, 4694--4702.

\bibitem[{Zhang et~al.(2020)Zhang, Zhu, Zhu, and Yang}]{zhang2020motion}
Zhang, H.; Zhu, L.; Zhu, Y.; and Yang, Y. 2020.
\newblock Motion-Excited Sampler: Video Adversarial Attack with Sparked Prior.
\newblock \emph{arXiv preprint arXiv:2003.07637}.

\bibitem[{Zhou et~al.(2016)Zhou, Khosla, Lapedriza, Oliva, and
  Torralba}]{zhou2016learning}
Zhou, B.; Khosla, A.; Lapedriza, A.; Oliva, A.; and Torralba, A. 2016.
\newblock Learning deep features for discriminative localization.
\newblock In \emph{Proceedings of the IEEE conference on computer vision and
  pattern recognition}, 2921--2929.

\bibitem[{Zhou et~al.(2018)Zhou, Hou, Chen, Tang, Huang, Gan, and
  Yang}]{zhou2018transferable}
Zhou, W.; Hou, X.; Chen, Y.; Tang, M.; Huang, X.; Gan, X.; and Yang, Y. 2018.
\newblock Transferable adversarial perturbations.
\newblock In \emph{Proceedings of the European Conference on Computer Vision
  (ECCV)}, 452--467.

\bibitem[{Zwillinger and Kokoska(1999)}]{zwillinger1999crc}
Zwillinger, D.; and Kokoska, S. 1999.
\newblock \emph{CRC standard probability and statistics tables and formulae}.
\newblock Crc Press.

\end{thebibliography}

\end{document}